**xML-workFlow: an end-to-end explainable scikit-learn workflow for rapid biomedical experimentation**

Khoa A. Tran[1#], John V. Pearson[2], Nicola Waddell[1]

[1]Medical Genomics Group, Cancer Program, QIMR Berghofer Medical Research Institute; Brisbane, Australia.

[2]Genome Informatics Group, Cancer Program, QIMR Berghofer Medical Research Institute; Brisbane, Australia.

# Corresponding author:

Dr Khoa A. Tran, Cancer Program, QIMR Berghofer Medical Research Institute, 300 Herston Road, Brisbane, 4006, Australia. Tel: +61 7 3845 0108. email: khoa.tran@qimrb.edu.au. ORCID: 0000-0001-7376-5061

**ABSTRACT**

**Motivation**

Building and iterating machine learning models is often a resource-intensive process. In biomedical research, scientific codebases can lack scalability and are not easily transferable to work beyond what they were intended. xML-workFlow addresses this issue by providing a rapid, robust, and traceable end-to-end workflow that can be adapted to any ML project with minimal code rewriting.

**Results**

We show a practical, end-to-end workflow that integrates scikit-learn, MLflow, and SHAP. This template significantly reduces the time and effort required to build and iterate on ML models, addressing the common challenges of scalability and reproducibility in biomedical research. Adapting our template may save bioinformaticians time in development and enables biomedical researchers to deploy ML projects.

**Availability and implementation**

xML-workFlow is available at https://github.com/MedicalGenomicsLab/xML-workFlow.

**INTRODUCTION**

Cancer is a complex and heterogeneous disease, necessitating the need for multi-omics datasets (e.g genomics, transcriptomics, proteomics, metabolomics) that are typically high-dimensional and may contain intricate interactions. Machine learning (ML) algorithms are capable of learning complex relationships in large amounts of data and have emerged as a powerful approach for uncovering patterns and making predictions from these large, complex datasets.[1].

Despite the promises of ML, the process of creating and iteratively refining ML models can be laborious and time-consuming. The complex nature of cancer biology may require numerous feature engineering strategies, algorithm choices, and hyperparameter configurations to optimise model performance to a specific biological question. This iterative and exploratory nature of biomedical ML research can result in technical challenges as a project progresses. This may include large codebases that require increasingly technical overheard to manage and scale as the number of experiments and datasets grows. This can lead to inefficient workflows, increased technical debt (i.e., managing old and/or repetitive code), and potential coding errors.

Building an ML pipeline typically requires four components: features engineering, training and validation, models and data management, and explainability (i.e., feature importance analysis). In biomedical research, features engineering is usually conducted using domain expert knowledge to curate biologically meaningful features[2-4]. Building the remaining three components of end-to-end ML can require substantial technical work. There have been efforts in creating scikit-learn templates[5-8]; however, these resources are not suitable for rapid experimentation as they were either case-specific[7], provided only in Jupyter notebook formats[5,8], or lacking models and data version management[5-8]. As a result, most scientific ML code is usually written to exist within the lifecycle of its related publication and may be challenging to transfer to other projects.

Here we present xML-workFlow, an explainable and scalable scikit-learn workflow designed specifically for biomedical researchers to rapidly scale up their ML experimental workflows. This technical note introduces xML-workFlow, which provides a streamlined and efficient framework for rapidly building and iterating over ML models, enabling researchers to scale up to dozens of experiments with ease. xML-workFlow integrates seamlessly with popular tools like MLflow for experiment tracking and includes built-in SHAP explainability components, empowering researchers to not only build accurate models but also understand and interpret their predictions within the context of their biological questions.

**IMPLEMENTATION**

xML-workFlow integrated three ML paradigms into an end-to-end workflow: models development - Train and Validation using scikit-learn[9], stateful models management - MLflow[10], and explainabilty - SHAP[11] (Figure 1). This workflow is designed to enable ML researchers to rapidly create new ML experiments with minimal technical overhead and allows tracking of experiment-specific models and data artefacts.

**Modularised machine learning workflow**

In the xML-workFlow workflow, we have modularised an ML project into three layers of abstraction (Figure 1). At the lowest level, we abstract constants (in *constance.py*, such as genome references or target variables) and parameters (in *params.py*, such as model configurations or scoring metrics) that are expectedly immutable across the entire experiment. The second and third layers of abstraction include mutable and stage-specific Python modules that typically evolve over the course of an experiment. In the second layer, we have modularised utilities code for training, validation, and explainability into *train_func.py*, *test_func.py*, and *shap_func.py*. The third layer provides wrapper code to execute each stage: *train_func.py → train.py*, *test_func.py → test.py*, and *shap_func.py → shap.py*. Of note, this separation between the second and third abstract layers was a deliberate engineering decision to facilitate code mutability. For example, *train_function.py* contains two modules: *prepare_train_data()* and *run_training()*, with the latter executing *GridSearchCV* or

*RandomizedSearchCV*. Any modifications to the *GridSearchCV / RandomizedSearchCV* training regime, which triggers the creation of a new sub-experiment, would only involve modifications to *train_func.py* while train.py is left unchanged and therefore reusable.

**Stateful model and data tracking with MLflow**

ML publications are typically released with the final version of their trained ML models and related training/validation code[2,3,12]. However, in practice biomedical researchers may create numerous experiments with various model and data versions. Without a tracking and management solution, this can result in reproducible errors, e.g., a published ML model version does not match its encompassing manuscript's descriptions. MLflow provides addresses this challenge by enabling stateful record-keeping (i.e., logging) for ML experimentation. The MLflow Python interface provides simple functions to log five essential aspects of an ML experiment:

1) training pipeline (usually enabled by the scikit-learn *Pipeline* API),
2) configurations of the ML model,
3) input and output descriptions,
4) snapshot of the data used for training,
5) k-fold cross-validation and post-training performance

xML-workFlow incorporates MLflow and automatically logs all five essential ML aspects described above by default throughout the whole workflow (Figure 1). Furthermore, MLflow is configured to save all information in a single location which is the project's home directory. By simply executing "*mlflow ui*" in this directory, researchers have access to an interactive MLflow dashboard to manage all historical models and their related data, as well as training and test performance metrics. To the best of our knowledge, xML-workFlow is the first ML workflow providing this seamless MLflow integration.

**Built-in explainability**

The ability of ML models to "explain" the most influential input features driving their predictions is essential to clinical translation of AI[1]. To this extent, the explainability method SHAP[11] can mathematically quantify the individual influence of each input feature on a trained ML model towards a unique input sample. The utility of SHAP in uncovering the key features within trained ML models has been extensively investigated and validated in biomedical research[4,13-16].

We adapted the methodology of calculating feature-specific median SHAP scores across test samples in one of our previous studies[16] and integrated it into xML-workFlow as an automated step. Therefore, once a trained ML model becomes available, xML-workFlow automatically:

1) applies the appropriate SHAP Explainer to the trained model across training samples
2) calculates median absolute SHAP scores for each feature across correctly predicted training samples, which should be all or nearly all of training samples if a model was fitted appropriately
3) generates interactive (*.html*) and static (*.png*) bar charts of the data in 2) and logs them as artifacts with MLflow
4) logs SHAP scores with corresponding model and data versions with MLflow
5) if test data is available, repeats step 2-4 across test samples.

**CONCLUSION**

In this technical note, we presented xML-workFlow, an end-to-end workflow that integrates scikit-learn[9], MLflow[10], and SHAP[11] into a single platform to enable rapid yet robust, scalable, and traceable ML experimentation.

Creating a new ML pipeline from scratch can take up to several weeks even for an experienced ML engineer. Moreover, scaling up ML experiments usually requires making the correct architectural decisions, which can only result from years of software engineering experience. xML-workFlow was designed to enable biomedical researchers to undertake ML projects.

We note that xML-workFlow is not a one-size-fits all solution. xML-workFlow could be further developed to support non-scikit-learn models, e.g., TensorFlow or Keras neural networks, which requires a completely different thinking paradigm. Additionally, while we created xML-workFlow as biomedical researchers and for biomedical researchers, it is our expectation that this workflow could be adapted in any domains that use scikit-learn to build ML models.

**DECLARATIONS**

**Ethics approval and consent to participate**

No patients were consented directly to this study. No participants were recruited for this study.

**Competing interests**

John V Pearson and Nicola Waddell are co-founders of genomiQa, a spin out company from QIMR Berghofer. The remaining authors declare that there are no competing interests.

**Acknowledgments**

This work and this research were performed on QIMR Berghofer computing infrastructure supported by the Australian Cancer Research Foundation (ACRF), The Ian Potter Foundation and The John Thomas Wilson Endowment. KAT was the recipient of the Maureen and Barry Stevenson PhD Scholarship; we are grateful to Maureen Stevenson for her support.

**Funding**

National Health and Medical Research Council of Australia (NHMRC) Investigator Grant 2018244 (NW)

Australian Cancer Research Foundation (ACRF) – The ACRF Centre for Optimised Cancer Therapy (JVP, NW)

**Author contributions**

Conceptualization: KAT, JVP, NW

Methodology: KAT, JVP, NW

Software: KAT

Formal analysis: KAT

Investigation: KAT

Supervision: JVP, NW

Project administration: JVP, NW

Funding acquisition: JVP, NW

Visualization: KAT

Writing – Original Draft: KAT

Writing - Review & Editing: JVP, NW

**REFERENCES**

1 Tran, K. A. *et al.* Deep learning in cancer diagnosis, prognosis and treatment selection. *Genome Med* **13**, 152 (2021). https://doi.org:10.1186/s13073-021-00968-x
2 Sammut, S. J. *et al.* Multi-omic machine learning predictor of breast cancer therapy response. *Nature* **601**, 623-629 (2022). https://doi.org:10.1038/s41586-021-04278-5
3 Crispin-Ortuzar, M. *et al.* Integrated radiogenomics models predict response to neoadjuvant chemotherapy in high grade serous ovarian cancer. *Nat Commun* **14**, 6756 (2023). https://doi.org:10.1038/s41467-023-41820-7
4 Tran, K. A. *et al.* Explainable machine learning identifies features and thresholds predictive of immunotherapy response. *bioRxiv*, 2025.2003.2023.643560 (2025). https://doi.org:10.1101/2025.03.23.643560
5 Yan, N. *handson-ml-implementation*, <https://github.com/hwyan0220/handson-ml-implementation> (2018).
6 Lapajne, J. *sklearn-project-template*, <https://github.com/Python-templates/sklearn-project-template> (2021).
7 Crispin-Ortuzar, M. *NAT-ML*, <https://github.com/micrisor/NAT-ML> (2021).
8 Lee, E. *A Beginner's Guide to Building Machine Learning Models with Scikit-learn: Step-by-Step Template*, <https://colab.research.google.com/drive/1XWFTmLSqZGubLMzM87kiYzfmkN0GhNQY?usp=sharing> (2024).
9 Pedregosa, F. *et al.* Scikit-learn: Machine Learning in Python. *The Journal of Machine Learning Research* **12**, 2825-2830 (2011).
10 Zaharia, M. A. *et al.* Accelerating the Machine Learning Lifecycle with MLflow. *IEEE Data Eng. Bull.* **41**, 39-45 (2018).
11 Lundberg, S. M. & Lee, S.-I. in *Proceedings of the 31st International Conference on Neural Information Processing Systems* 4768–4777 (Curran Associates Inc., Long Beach, California, USA, 2017).
12 Osipov, A. *et al.* The Molecular Twin artificial-intelligence platform integrates multi-omic data to predict outcomes for pancreatic adenocarcinoma patients. *Nat Cancer* **5**, 299-314 (2024). https://doi.org:10.1038/s43018-023-00697-7
13 Lee, S. I. *et al.* A machine learning approach to integrate big data for precision medicine in acute myeloid leukemia. *Nat Commun* **9**, 42 (2018). https://doi.org:10.1038/s41467-017-02465-5
14 Lundberg, S. M. *et al.* Explainable machine-learning predictions for the prevention of hypoxaemia during surgery. *Nat Biomed Eng* **2**, 749-760 (2018). https://doi.org:10.1038/s41551-018-0304-0
15 Chen, H., Lundberg, S. M., Erion, G., Kim, J. H. & Lee, S. I. Forecasting adverse surgical events using self-supervised transfer learning for physiological signals. *NPJ Digit Med* **4**, 167 (2021). https://doi.org:10.1038/s41746-021-00536-y
16 Yap, M. *et al.* Verifying explainability of a deep learning tissue classifier trained on RNA-seq data. *Sci Rep* **11**, 2641 (2021). https://doi.org:10.1038/s41598-021-81773-9

# FIGURES

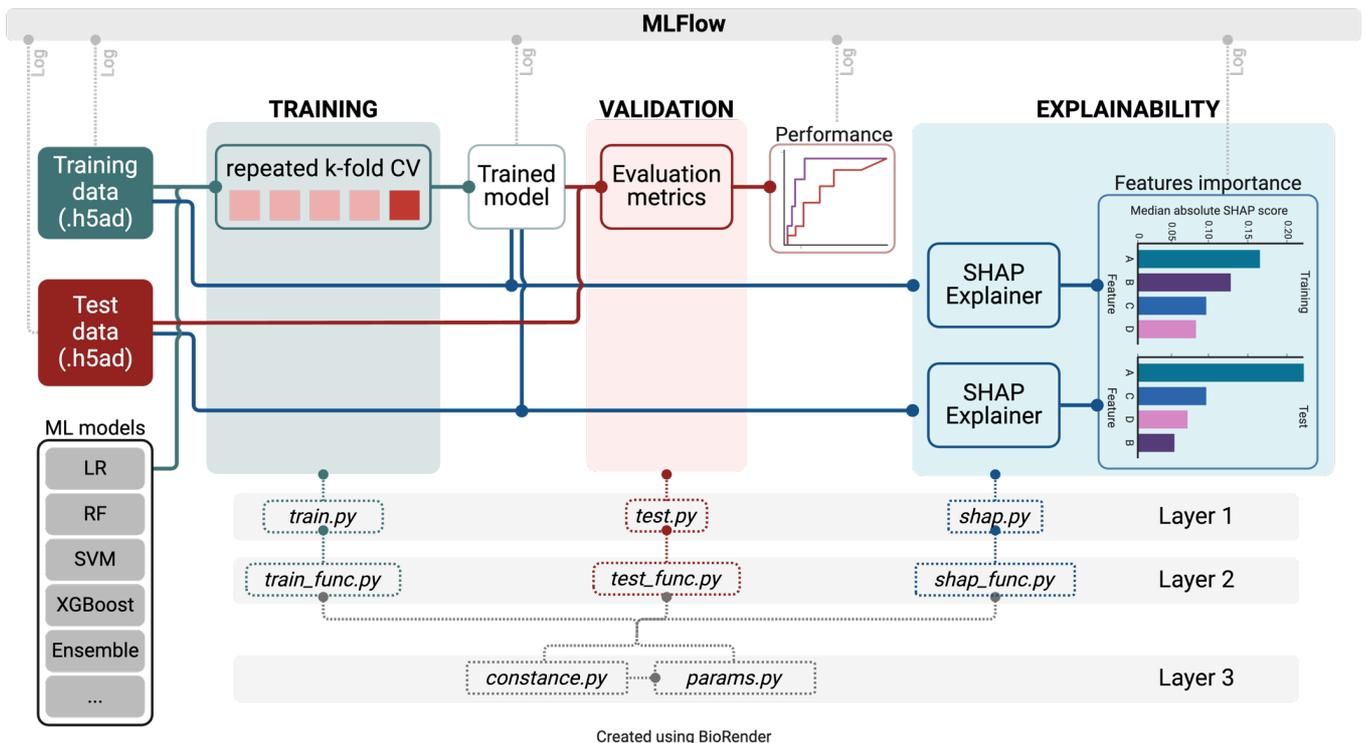

**Figure 1**. *xML-workFlow: An explainable and scalable scikit-learn machine learning workflow.* Visual illustration of xML-workFlow. Solid coloured boxes (green, red, and dark grey) represent input artefacts, such as machine learning models and datasets. Connected lines with rounded ends represent data flow and dependencies, e.g., methods in *test_func.py* are imported into *test.py*. Unfilled boxes with dark outlines represent computational steps within the workflow. Unfilled boxes with blurred colour lines represent output artefacts. Dotted lines connecting to MLFlow represent objects automatically tracked by MLFlow. LR: logistic regression, RF: random forest, SVM: support vector machine, XGBoost: eXtreme Gradient Boosting, SHAP: SHapley Additive exPlanations.